\documentclass[runningheads]{llncs}
\usepackage{graphicx}
\usepackage{enumitem}
\usepackage{mwe} 
\usepackage{cite}
\usepackage{amsmath,amssymb,amsfonts}
\usepackage{algorithmic}
\usepackage{textcomp}
\usepackage[table,xcdraw]{xcolor}
\usepackage{booktabs}
\usepackage{multirow}
\usepackage{bbding}
\usepackage{times}
\usepackage{epsfig}
\usepackage{soul}
\usepackage{mathtools}
\usepackage{dsfont}
\usepackage{tabularx}
\usepackage{wrapfig}
\usepackage{hyperref}
\definecolor{firebrick}{rgb}{0.7, 0.13, 0.13}

\hypersetup{
    colorlinks=true,
    linkcolor=blue,
    filecolor=magenta,      
    urlcolor=magenta,
    pdfpagemode=FullScreen,
    }

\begin{document}
\title{Simultaneous Semantic and Instance Segmentation \\ for Colon Nuclei Identification and Counting}
\titlerunning{Simultaneous Semantic and Instance Segmentation}
%
\author{Lihao Liu\inst{1} \and Chenyang Hong\inst{2} \and \\ Angelica I. Aviles-Rivero\inst{1} \and Carola-Bibiane Schönlieb\inst{1}}

\authorrunning{L. Liu et al.}
%
\institute{DAMTP, University of Cambridge \and
CSE, Chinese University of Hong Kong}
%

\maketitle              

\begin{abstract}
Nucleus segmentation and classification within the Haematoxylin and  Eosin stained histology images is a key component in computer-aided image analysis, which helps to extract features with rich information for cellular estimation and following diagnosis. Therefore, it is of great relevance for several downstream computational pathology applications. 
In this work, we address the problem of automatic nuclear segmentation and classification. Our solution is to cast as a simultaneous semantic and instance segmentation framework, and it is part of the 
the Colon Nuclei Identification and Counting (CoNIC) Challenge. Our framework is a  carefully designed ensemble model.
We first train a semantic and an instance segmentation model separately, where we use as backbone HoverNet and Cascade Mask-RCNN models. 
We then ensemble the results with a customized Non-Maximum Suppression embedding algorithm. 
From our experiments, we observe that the semantic segmentation part can achieve an accurate class prediction for the cells whilst the instance information provides a refined segmentation. We enforce a robust segmentation and classification result through our customized embedding algorithm.
%
%
We demonstrate, through our visual and numerical experimental,  that our model outperforms the provided baselines by a large margin.  Our solution ranked within the top 5 solutions on the Grand Challenge CoNIC~2022.
\end{abstract}

\begin{keywords}
Semantic Segmentation, Instance Segmentation, Histology Images, Colon Nuclei Identification and Counting
\end{keywords}

\section{Introduction}

Computational pathology plays an important role in accurate cellular estimation and also in diagnosis.
Specifically, by extracting rich and interpretable cell-based features, nuclei segmentation and classification are becoming a significant asset in the downstream explainable models in computational pathology.
Researchers have studied automatic methods for colon nuclei segmentation and classification in Haematoxylin and Eosin stained histology images for
decades. 
In recent years, the advent of convolutional neural networks (CNN) has revolutionised this field yielding state-of-the-art segmentation and classification results.
Although promising results have been reported, this task is still challenging due to the large difference in shape between different types of cells.
Traditional methods perform cell segmentation based on hand-crafted features e.g.~\cite{cheng2008segmentation,jung2010segmenting}.
%
However, due to the lack of highly semantic information, i.e. a global understanding of the images, traditional methods are still limited in performance.
%

Most recently, CNN-based methods~\cite{liu2018mtmr,liu2019multi, liu2020contrastive,liu2019probabilistic}, have demonstrated remarkable segmentation and classification accuracy. 
By extracting highly semantic features,  CNN-based methods are capable of learning a strong and robust representation of the images, hence, achieving better performance than traditional methods.
In particular, existing CNN-based segmentation techniques can be mainly categorised into semantic segmentation and instance segmentation.
%
Semantic segmentation focus on giving the image contents different classes. By contrast, instance segmentation 
treats multiple objects of the same class as distinct entities.

The body of literature on semantic segmentation has reported several potential methods, where the most widely used model is  U-Net~\cite{ronneberger2015u}. That model uses an encoder-decoder architecture to reconstruct segmentation masks from CNN features.
Based on U-Net, $\Psi$-Net~\cite{liu2020psi} achieve better senmantic segmentation accuracy bu stacking densely convolutional LSTMs.
%
Another well-known model is Hover-Net~\cite{graham2019hover}. It further leverages the instance information encoded within the vertical and horizontal distances of nuclear pixels to their centers of mass to perform instance segmentation.
%

%
Unlike semantic models, instance techniques usually utilise a two-stage architecture to detect and segment each instance.
%
A well-established technique is Faster RCNN~\cite{ren2015faster}. It is composed of a region proposal stage and a bounding box regression stage, which is effective for detection tasks.
Mask-RCNN~\cite{he2017mask} extends Faster R-CNN by adding a mask prediction branch, in parallel to the bounding box recognition branch, to perform instance detection and segmentation simultaneously.
Cascade Mask R-CNN~\cite{cai2018cascade} achieves better segmentation and classification performance than previous models. By adding a sequence of detectors trained with increasing IoU thresholds on Mask-RCNN. 

\textbf{Contributions.} 
From our observations, semantic models are able to do highly accurate class predictions for cells; as these types of techniques provide a global understanding of the image content. Whilst instance models provide a refined segmentation as they consider the segmentation on each instance.
%
Motivated by the aforementioned advantages of each family of techniques,  we propose a  simultaneous semantic and instance segmentation framework for Colon Nuclei identification and Counting.
%
Particularly, we firstly train a semantic and instance segmentation model
separately. We use as backbone the HoverNet and the Cascade Mask-RCNN. Secondly, we ensemble   the results with  a customized Non-Maximum
Suppression embedding algorithm. The output of our framework is the colon nuclei identification and counting. Our framework ranked within the top 5 solutions out of 373 submissions on the Grand Challenge CoNIC 2022.
%



\begin{figure}[t]
 \centering
 \includegraphics[width=\linewidth]{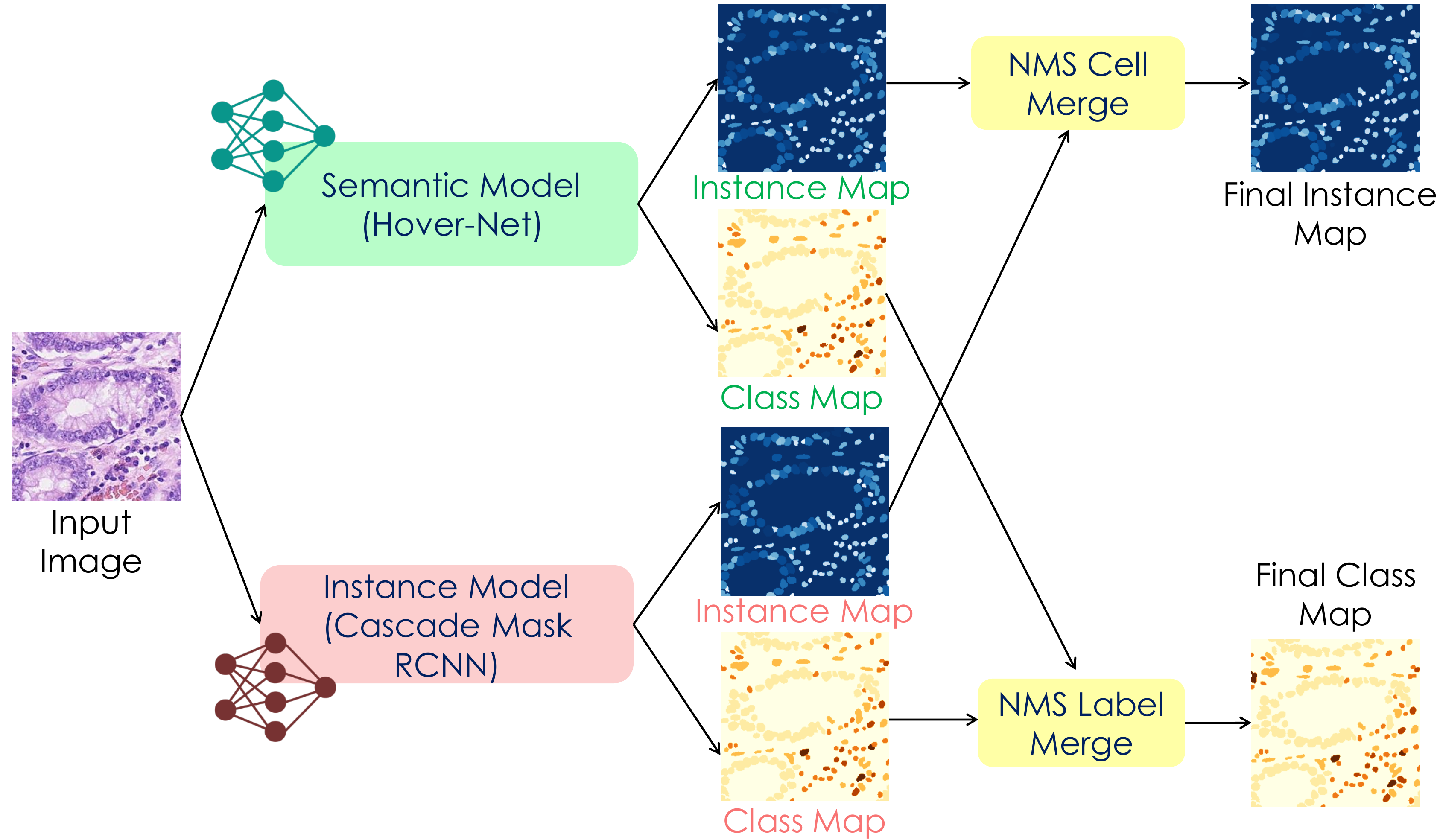} 
 \centering
 \caption{The schematic illustration of the proposed simultaneous semantic and instance segmentation framework. We first train a semantic and instance segmentation models separately. We then ensemble the results via our customised Non-Maximum Suppression embedding algorithm.}
 \label{fig:network}
\end{figure}

\section{Proposed Framework}
Our framework is composed of three key parts (see Fig.~\ref{fig:network}). i) an instance segmentation model, ii) a semantic segmentation model and iii) a customised Non-Maximum
Suppression (NMS) embedding algorithm.  We describe in detail these parts next.

\subsection{Every Single Nucleus Matters: Instance Segmentation Model}
Instance segmentation~\cite{ren2015faster,he2017mask} seeks to detect and localise objects from a set of predefined classes in an image. Unlike semantic segmentation, instance segmentation treats the objects of the same class as different instances. One of the most well-established models is the so-called Cascaded Mask-RCNN~\cite{cai2018cascade}, where a multi-stage object detection model is proposed. Because of the simplicity yet the high performance offered by Cascade R-CNN, we select it as our backbone.

%
%
%

We follow the work of that~\cite{cai2018cascade,liu2020psi}, where we seek to optimise the following the element-wise cross entropy loss over all voxels and all categories:
\begin{equation} \label{loss_seg}
\mathcal{L}_{instance} = -\frac{1}{M\times K}\sum_{m=1}^M \sum_{k=1}^{K} \ y_m^k \ \log(p_m^k) \ ,
\end{equation}
where $M$ denotes  the total number of pixels in an input image, and
$K$ is the cell categories in our application we have seven categories including the background (details on the categories are displayed in the experimental section).
Moreover, $y_m^k$ refers to the ground truth label at the $m$-th pixel in the $k$-th category; $p_m^k$ is the softmax output value, which represents a predicted probability indicating the $m$-th voxel belongs to the $k$-th category.

\begin{figure}[t]
 \centering
 \includegraphics[width=\linewidth]{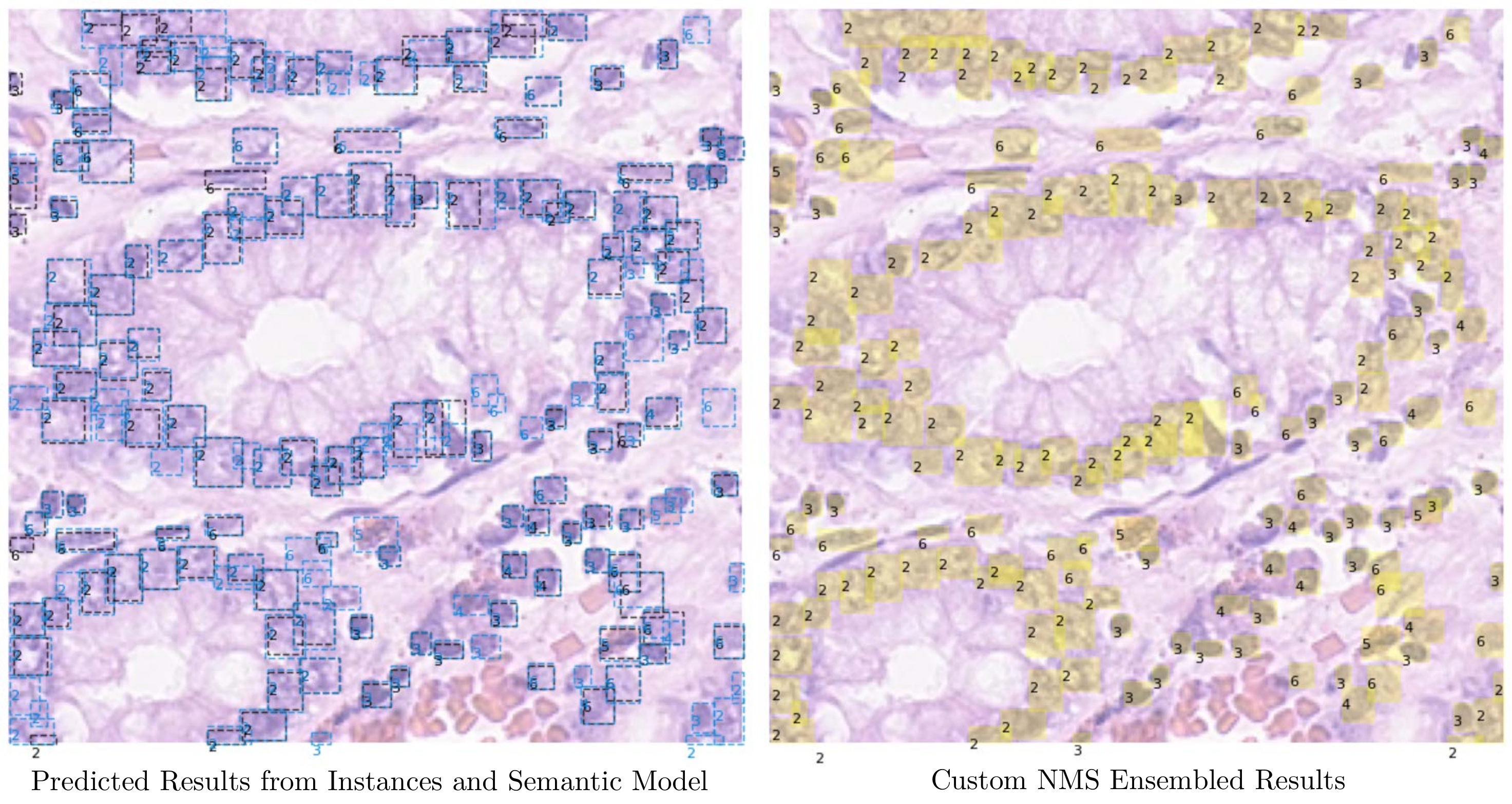} 
 \caption{Illustration of the ensemble process from the NMS. The left side figure displays the predicted bounding boxes from the semantic model (black bounding boxes) and the instance model (blue bounding boxes), respectively. We then ensemble the results using our customised NMS algorithm. 
 The final output is displayed, on the right side, showing the ensemble results visualised on the yellow bounding boxes.}
 \label{fig:nms}
\end{figure}

We provide the detailed setting of the Cascaded Mask-RCNN for our purpose.
To extract meaningful feature maps, we select ResNeXt-152~\cite{cai2018cascade} as the backbone for the Cascaded Mask-RCNN.
For natural images, 
instance segmentation models usually take a large-sized image as input, e.g. 1300 $\times$ 800. 
If the image is too small, then the segmentation is coarse.
To meet this specification, 
we set the input as 512 $\times$ 512, which is currently the smallest image size for instance segmentation models.
Moreover, we decrease the original anchor size setting from (32, 64, 128, 256, 512) to (8, 16, 32, 64, 128) due to the smaller size of the images and their targets (nucleus).
With these changes, the model is more robust when detecting smaller objects like the nucleus. We display a sample output of the instance model on the left side of Fig.~\ref{fig:nms}. Most notable, the output is marked using the blue bounding boxes.

\subsection{Denser Predictions for Better Performance: Semantic Segmentation Model}
Semantic segmentation is to provide a denser prediction than instance segmentation models. As the goal is to predict pixel-wise the label of objects related to a set of defined classes.
To apply a semantic segmentation model on instance segmentation tasks, an additional post-processing step is needed.
This post-processing process can split the semantic segmentation result into multiple single objects from the same class (i.e. into instance segmentation). 
%
To avoid re-implement this function, we directly select Hover-Net as a semantic training model. That is, we seek to optimise another loss, which definition follows~\eqref{loss_seg}. For clarity purposes, we denote the loss as $\mathcal{L}_{semantic}$.
We can therefore focus on the model design rather than the post-processing.

To keep the same setting as the instance segmentation model, we select ResNeXt-152~\cite{xie2017aggregated} as the backbone for our semantic segmentation model, and we use the same input size 512 $\times$ 512 for consistency. 
We remark that we only did these two modifications, and the rest of the network remains unchanged following Hover-Net~\cite{graham2019hover}. 
We display a sample output of the semantic model on the left side of Fig.~\ref{fig:nms}. The output is highlighted using the black bounding boxes.

%

\subsection{Model Ensemble: A Non-Maximum Suppression Embedding Algorithm}
After obtaining the results from the semantic and instance segmentation models, we then adopted a custom NMS model to ensemble the detected instances from the two models.
We remark that this step is not a learning process.
NMS seeks to find overlapped instance results based on the position of the detected cells bounding boxes.
By setting an Intersection over Union (IoU) threshold, it can be determined if two or more cells from the two models identify the same cell.
We highlight that the standard NMS only eliminates multiple excessive cells and keep only one cell.
In this work, \textit{we proposed a customised} NMS, which not only detects multiple cells but also merges the segmentation mask and the classification labels.

We first detected the overlapped segmentation results for one cell / one position. We then take as a final segmentation output the merge of such overlapped regions of a cell. 
%
Then, we assign a classification label to the merged segmentation results by a weighted voting mechanism, which details are as follows.
The semantic segmentation model is capable of a great understanding of the global information content. Hence, it can achieve a great classification performance in cells with a small number such as neutrophils, plasma, and eosinophils.
For the label merging process, the weight of the neutrophil, epithelial, lymphocyte, plasma, eosinophil, and connective tissue cells is taken from the semantic segmentation model, and set to (2, 1, 1, 2, 2, 1).
Whilst the weight of neutrophil, epithelial, lymphocyte, plasma, eosinophil, and connective tissue cells is taken from the instance segmentation model and set to (1.5, 1.5, 1.5, 1.5, 1.5, 1.5). 
We assign 1.5 to the weight of the instance model's prediction to avoid the same voting number.
For example, two cells overlapped in the NMS stage, where the semantic model predict it as neutrophil and the instance model as epithelial. Then the final label for this cell is assigned as a neutrophil.
We underline that we did not use any other further post-processing techniques to refine the results. On the right side of Fig.~\ref{fig:nms}, we display a visual sample output from the ensemble model.


\section{Experimental Results}
In this section, we detail all experiments and conditions that we follow to evaluate our framework.

\begin{wrapfigure}{r}{0.48\textwidth} \vspace{-1.2cm}
  \begin{center}
    \includegraphics[width=0.48\textwidth]{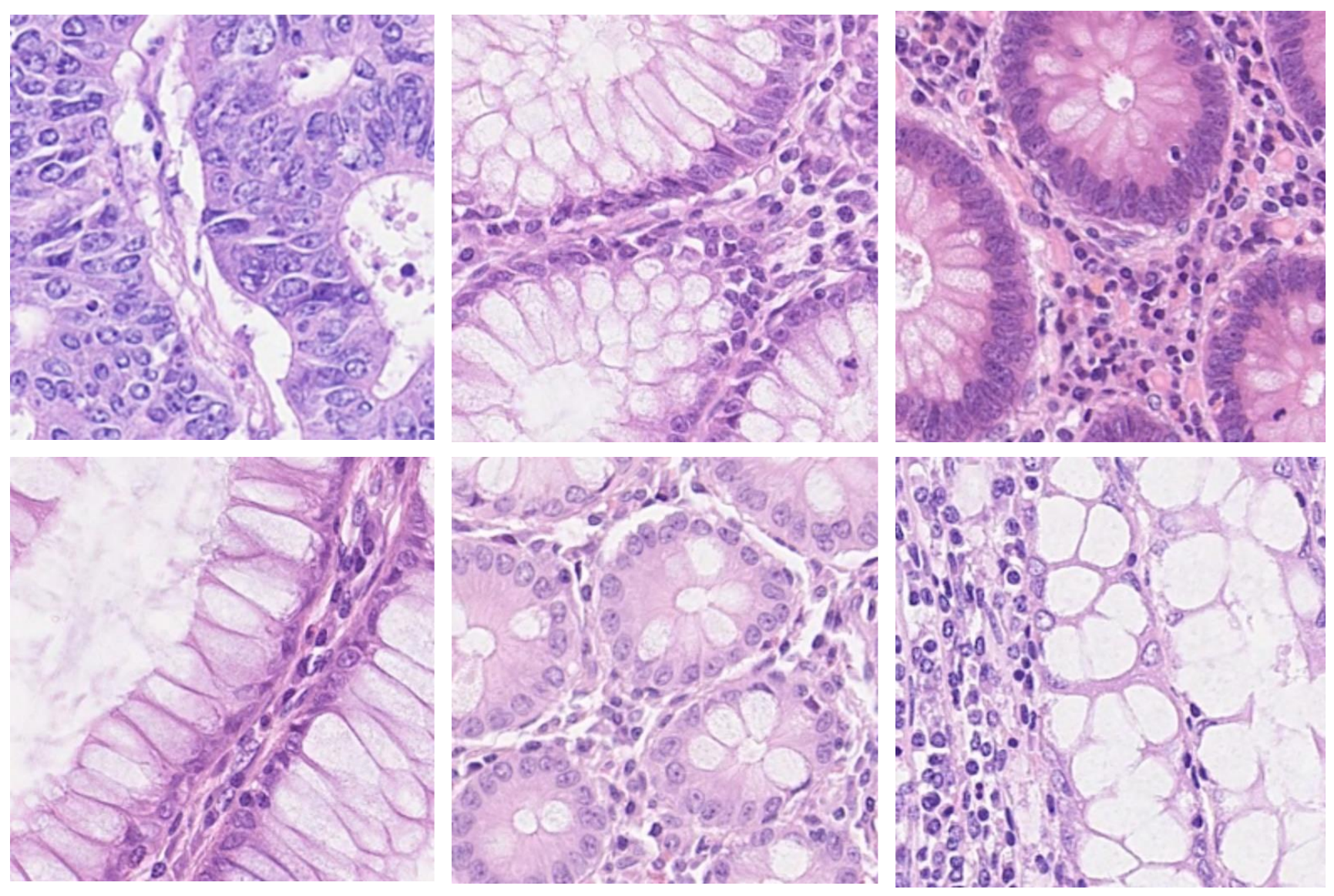} 
  \end{center}
  \vspace{-0.6cm}
  \caption{Visual samples of the histology images used in our experiments. We display selected samples to  show the diversity and complexity of the dataset.} \label{fig:samples} \vspace{-0.5cm}
\end{wrapfigure}

\medskip
\textbf{Challenge Dataset Description.}
The Colon Nuclei Identification and Counting (CoNIC) Challenge~\cite{graham2021conic} dataset is forked from the Lizard dataset~\cite{graham2021lizard}, which is the largest known publicly available dataset in the computational pathology area for instance segmentation.
The dataset is composed of Haematoxylin \& Eosin stained histology images from six different data sources, with a unified cropped size of 256 $\times$ 256. Visual samples of the dataset are displayed in Fig.~\ref{fig:samples}.
For each target nucleus, an instance mask and a classification mask are provided, where each nucleus is assigned to one of the following categories: 1--Neutrophil, 2--Epithelial, 3--Lymphocyte, 4--Plasma, 5--Eosinophil, and 6--Connective tissue. 

\medskip
\textbf{Pre-processing and Data Augmentation}
For a robust training process, we use a four steps augmentation process to enrich the training data.
[Step 1] We firstly enlarged the image to twice its original size 512 $\times$ 512. This step is necessary for our instance model training as we described in Section 2.
[Step 2] We adopted a random scale to resize the images to 0.8 - 1.2 of the size drawn from Step 1. We then randomly cropped, using a fixed size of 512$\times$512, regions with zero padding. 
[Step 3] We used random flip to augment the data from different views.
[Step 4] We randomly selected Gaussian blur, median blur, and additive Gaussian noise. We used the imgaug package~\footnote{\url{https://github.com/aleju/imgaug}}
to enhance the pixels sensitivity of the input. 

\medskip
\textbf{Implementation Details.}
Our code is built based on Hover-Net and detectron2 repositories.
For the loss function, we use binary cross-entropy loss for mask prediction, and cross-entropy loss for label prediction in both repositories.
For the instance model, SGD optimiser is used during training with the initial learning rate setting as $10^{-4}$. Whilst, for the semantic model, Adam optimiser is used with the same learning rate $10^{-4}$.
All models are run on an NVIDIA A100 GPU with 80G RAM, which takes around 12 hours to train the instance model for 60 epochs, and 96 hours to train the semantic models for 100 epochs.
{To evaluate our model, we first use the multi-class panoptic quality (PQ) and multi-class PQ$^+$ metrics to evaluate the segmentation and classification tasks. Whilst for the counting (cellular composition) the multi-class coefficient of determination R$^2$.}

\begin{table}[t]
\centering
\resizebox{1\textwidth}{!}{
\begin{tabular}{c|c|ccc|cccccc} \toprule[1pt]
    \cellcolor[HTML]{EFEFEF}Model 
    & \cellcolor[HTML]{EFEFEF}DataAug 
    & \cellcolor[HTML]{EFEFEF}R$^2$  
    & \cellcolor[HTML]{EFEFEF}PQ    
    & \cellcolor[HTML]{EFEFEF}mPQ$^{+}$ 
    & \cellcolor[HTML]{EFEFEF}PQ$^{+}_{neu}$ 
    & \cellcolor[HTML]{EFEFEF}PQ$^{+}_{epi}$
    & \cellcolor[HTML]{EFEFEF}PQ$^{+}_{lym}$ 
    & \cellcolor[HTML]{EFEFEF}PQ$^{+}_{pla}$
    & \cellcolor[HTML]{EFEFEF}PQ$^{+}_{eos}$
    & \cellcolor[HTML]{EFEFEF}PQ$^{+}_{con}$  \\
    \midrule[0.8pt] 
    
     {B}    & {\XSolidBrush}  & 0.8585              & 0.6149	            & 0.4998             & 0.2435             & 0.6384             & 0.6832              & 0.5066              & 0.3946              & 0.5707             \\ \midrule[0.8pt]
     {S}    & {\XSolidBrush}  & 0.8697              & 0.6235                & 0.5195             & 0.3024             & 0.6133             & 0.7079              & 0.5136              & 0.3954              & 0.5843             \\
     {S}    & {\Checkmark}    & \underline{\textcolor{firebrick}{0.8801}}  & 0.6310                & \underline{\textcolor{firebrick}{0.5338}} & \underline{\textcolor{firebrick}{0.3498}} & 0.6011             & \underline{\textcolor{firebrick}{0.7233}} & \underline{ \textcolor{firebrick}{0.5326}}  & \underline{\textcolor{firebrick}{0.4007}}  & 0.5955             \\
     {I}    & {\XSolidBrush}  & 0.7933              & 0.6369                & 0.5158             & 0.2858             & 0.6421             & 0.6955              & 0.5076              & 0.3546              & 0.6090             \\ 
     {I}    & {\Checkmark}    & 0.8020              & \underline{\textcolor{firebrick}{0.6457}}    & 0.5294             & 0.3208             & \underline{\textcolor{firebrick}{0.6491}} & 0.7070              & 0.5073              & 0.3750              & \underline{\textcolor{firebrick}{0.6167}} \\ \midrule[0.8pt]
     {Ours} & {\Checkmark}    &\textbf{\textcolor{blue}{0.9048}}      & \textbf{\textcolor{blue}{0.6658}}       & \textbf{\textcolor{blue}{0.5655}}    & \textbf{\textcolor{blue}{0.3781}}    & \textbf{\textcolor{blue}{0.6225}}    & \textbf{\textcolor{blue}{0.7373}}     & \textbf{\textcolor{blue}{0.5621}}     & \textbf{\textcolor{blue}{0.4293}}     & \textbf{\textcolor{blue}{0.6272}}    \\ \midrule[0.8pt]
\end{tabular}
}
\caption{Experimental result on the discovery phase. We refer to 'S' and 'I' as the semantic and instance models respectively, and  'B' as the baseline provided by the challenge organizers. 'DataAug' refers to data augmentation. The \textcolor{blue}{\textbf{best results}} are highlighted in bold and blue colour. Whilst the \textcolor{firebrick}{second best} is denoted in orange colour. } 
\label{results}
\end{table}

\begin{figure*}[t]
 \centering
 \includegraphics[width=\textwidth]{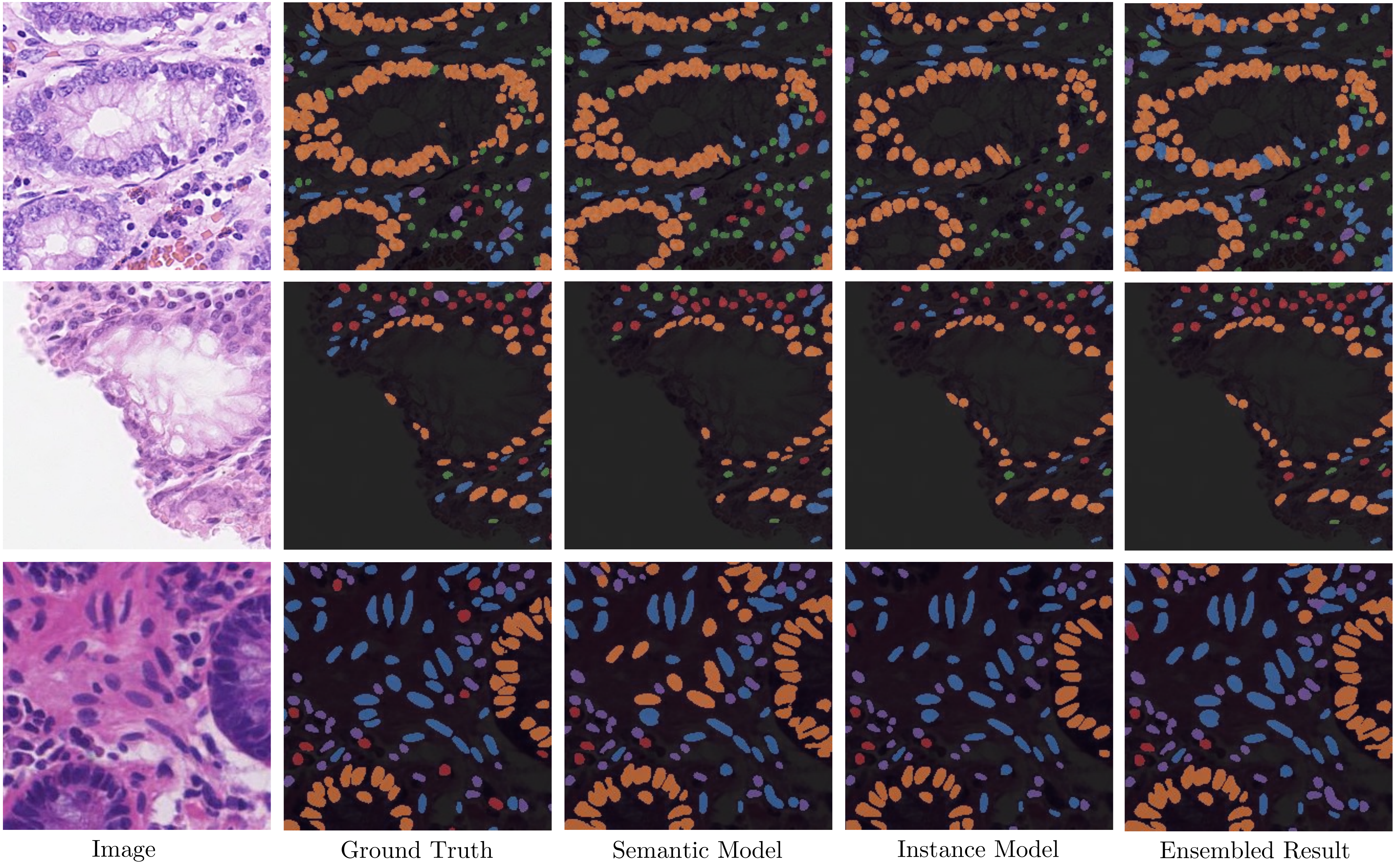} 
 \centering
 \caption{Visual comparison of the outputs produced by the Semantic Model, Instance Model, and our Proposed Ensemble framework.  The color code represents different classes that the cell belongs to (e.g. Neutrophil and Plasma).
}
 \label{fig:results}
\end{figure*}

\subsection{Results \& Discussion}
%
We followed the official data splitting, which uses 3900 images as the training set and 1081 images as the testing set.
We compare our ensembled results against: i) the Baseline and Semantic Model (we modified the backbone and input size) provided by the CoNIC challenge~\cite{graham2021conic}, and ii) Detectron2\footnote{\url{{https://github.com/facebookresearch/detectron2}}} (Instance Model).

The results are reported in~Table~\ref{results}, where our framework outperformed all compared models. In a closer inspection, we observe that our modified Semantic Model performs better than the baseline model. 
%
The performance improvement comes from the data augmentation and better backbone model.
Moreover, one interesting finding is that the Instance Model performs better than the Semantic Model in terms of PQ evaluation metrics, whilst the Semantic Model outperforms the instances model in terms of mPQ+ and R$^2$ evaluation metrics. 
This is because, when performing segmentation, the Instance Model focuses only on the nucleus region. Hence, it has a better segmentation accuracy (PQ).
While the Semantic Model has a great semantic understanding of the content of the image, i.e. how many cells are in the images. Hence, the classification accuracy (R$^2$) is better. 
Also, even with a low PQ, the Semantic Model can also outperform the Instance Model regarding the final mPQ+.

Our ensembled model (Ours) deals with the inherent problems from the baseline and the Semantic and Instance Models. By providing a good trade-off between the image content and the nucleus region information. As consequence, our framework outperforms all compared models for all metrics.
We highlight the positive performance impact of 
combining both semantic and instance information, see the visual result of our NMS ensemble in Fig.~\ref{fig:nms}. We further support our numerical results with a set of visual comparisons (see Fig.~\ref{fig:results}) reflecting the advantage of our framework.

\textbf{Test Stage Performance.} There are three main phases for the CoNIC Challenge: Discovery phase, Preliminary test phase, and Final test phase.
The results from Table~\ref{results} are from the discovery phase. At the time of submitting this paper, we had the test phase results, which details are as follows.
In the preliminary testing stage, we use 5-fold cross-validation, and trained 5 models. We ensemble the 5 models and submit them using docker.
However, due to the limited testing time, we are only able to submit the instance model, since currently, the semantic model is taking longer than expected to test (see Fig.~\ref{fig:network}).
Our instance model has reached an mPQ+ of 0.47540 on the preliminary testing dataset for the segmentation task, and an R$^2$ of 0.67599 on the preliminary testing dataset for the counting task.

\section{Conclusion} 
In this paper, we introduce a framework for colon nuclei identification and counting. Our solution is framed as a simultaneous semantic and instance segmentation framework. Our carefully designed solution is an ensemble that considers simultaneously the nucleus regions and the semantics of the images; this ensures focus on relevant regions on the image content and therefore boosts the performance. Our ensemble is achieved via a customised Non-Maximum Suppression embedding algorithm. Our results outperformed the challenge baseline. We highlighted the relevance of considering semantic and instance information for the task at hand. We underline that our framework ranked within the top 5 solutions out of 373 submissions on the Grand Challenge CoNIC 2022.
%
%

%
%
\bibliographystyle{splncs04}
\bibliography{refs}
\end{document}